\documentclass{article}

\usepackage{microtype}
\usepackage{graphicx}
\usepackage{subcaption}
\usepackage{booktabs}
\usepackage{siunitx}
\usepackage{multirow}
\usepackage{float}
\usepackage{tabularx}
\usepackage{makecell}
\usepackage{amsmath}
\usepackage{amssymb}
\usepackage{mathtools}
\usepackage{amsthm}
\usepackage{hyperref}
\usepackage[capitalize,noabbrev]{cleveref}

\usepackage{hyperref}

\usepackage[preprint]{icml2026}
\usepackage{amsmath}
\usepackage{amssymb}
\usepackage{mathtools}
\usepackage{amsthm}

\usepackage[capitalize,noabbrev]{cleveref}

\theoremstyle{plain}
\newtheorem{theorem}{Theorem}[section]

\newtheorem{lemma}[theorem]{Lemma}

\theoremstyle{definition}

\theoremstyle{remark}

\usepackage[textsize=tiny]{todonotes}

\icmltitlerunning{Submission and Formatting Instructions for ICML 2026}

\begin{document}

\twocolumn[
  \icmltitle{SDSC: Signal Dice Coefficient \\ A Structure-Aware Metric for Semantic Signal Representation Learning}




  \begin{icmlauthorlist}
    \icmlauthor{Jeyoung Lee}{yyy,comp}
    \icmlauthor{Hochul Kang}{yyy}
  \end{icmlauthorlist}
  
\icmlaffiliation{yyy}{Department of Digital Media Engineering, The Catholic University of Korea, Bucheon-si Gyeonggi-do, Republic of Korea}
\icmlaffiliation{comp}{Sweetndata, Seoul-si,  Republic of Korea}

\icmlcorrespondingauthor{Hochul Kang}{hckang19@catholic.ac.kr}

  \icmlkeywords{Machine Learning, ICML}

  \vskip 0.3in
]



\printAffiliationsAndNotice{}  

\begin{abstract}
We introduce the Signal Dice Similarity Coefficient (SDSC), a structure-aware objective for self-supervised representation learning on time-series data. Most existing reconstruction-based methods rely on distance-based losses such as mean squared error (MSE), which are sensitive to amplitude scale, largely invariant to waveform polarity, and unbounded in value. These properties can obscure structural fidelity and complicate interpretation, particularly when reconstruction error does not reflect semantic consistency.

SDSC measures local structural agreement through signed amplitude intersections, extending the Dice Similarity Coefficient from segmentation to continuous, signed signals. Although SDSC is defined as a similarity metric, it can be incorporated as a training objective by minimizing $1-\mathrm{SDSC}$ using a differentiable approximation of the Heaviside function. To address amplitude-critical scenarios and improve optimization stability, we further propose a hybrid objective that combines SDSC with MSE.

Across forecasting and classification benchmarks, SDSC-based pretraining yields downstream performance comparable to, and in some cases more consistent than, MSE-based objectives under identical contrastive settings. In particular, SDSC demonstrates stable behavior in in-domain and low-resource regimes, where preserving local waveform structure is beneficial. These results suggest that explicitly encouraging structural consistency in reconstruction complements distance-based objectives and motivates structure-aware alternatives for time-series self-supervised learning.
\end{abstract}

\section{Introduction}

Self-supervised learning (SSL) enables representation learning from unlabeled data by optimizing proxy objectives such as masked prediction or signal reconstruction. In domains such as computer vision and natural language processing, SSL has demonstrated strong capability in extracting semantically meaningful representations that transfer effectively to downstream tasks~\citep{ssl_survey}
In time-series modeling, SSL has similarly been adopted to learn representations for downstream tasks including forecasting and classification~\citep{signal_survey}. In many real-world signals, such as EEG or EMG, task-relevant semantics are often reflected in structural properties of the waveform, including local shape patterns, polarity changes, and characteristic temporal variations. Capturing these signal-specific structures is therefore an important aspect of representation learning. However, most reconstruction-based SSL methods for time-series data rely on distance-based losses, most commonly mean squared error (MSE), which emphasize pointwise amplitude differences.
While effective for minimizing numerical reconstruction error, distance-based metrics exhibit several limitations when used as objectives for semantic representation learning. First, they are highly sensitive to amplitude scale and largely invariant to waveform polarity, allowing reconstructions that minimize error without preserving meaningful structural consistency. For example, phase-inverted signals, amplitude-scaled signals, or near-zero baselines can yield similar MSE values despite representing substantially different semantics. Second, distance-based losses are unbounded and lack normalization, which complicates interpretation and model selection. As a result, low reconstruction error does not necessarily correspond to high semantic fidelity. Recent studies have suggested that the default reliance on MSE may not be optimal for all time-series learning scenarios~\citep{zeng2023transformers}.
To address these limitations, we introduce the Signal Dice Similarity Coefficient (SDSC), a structure-aware metric designed to quantify local structural agreement between temporal signals. SDSC is inspired by the Dice Similarity Coefficient (DSC)~\citep{dice_score,dice_score2}, which is widely used in semantic segmentation, and extends it to continuous, signed time-series signals. The proposed metric is bounded within $[0,1]$, exhibits reduced sensitivity to amplitude variation, and explicitly reflects agreement in local waveform structure. Although SDSC is defined as a similarity metric, it can be incorporated as a training objective by minimizing $1-\mathrm{SDSC}$. To enable gradient-based optimization, we adopt a differentiable approximation of the Heaviside function. In addition, we propose a hybrid reconstruction objective that combines SDSC with MSE to improve optimization stability while preserving amplitude information when necessary.
Self-supervised learning frameworks for time-series data typically consist of two complementary components: reconstruction-based objectives and contrastive learning. While contrastive losses explicitly encourage instance-level discrimination, reconstruction objectives influence how signal content and structure are encoded in the latent representation. In this context, the choice of reconstruction loss plays a critical role in shaping the learned representations. Despite their widespread use, distance-based losses primarily reflect amplitude fidelity and may implicitly capture structure only as a byproduct. The observation that MSE- and SDSC-based models often achieve comparable downstream performance suggests that low reconstruction error alone does not guarantee faithful semantic preservation.
In this study, SDSC is integrated exclusively into the reconstruction branch of SimMTM, a self-supervised framework that combines reconstruction and contrastive learning. The contrastive objective (InfoNCE)~\citep{nce} is kept identical to the original SimMTM formulation, allowing us to isolate the effect of the reconstruction objective under controlled conditions. Throughout this paper, the term \emph{structure-aware} specifically refers to local waveform similarity characterized by pointwise sign agreement and magnitude overlap. SDSC is therefore alignment-free and computationally linear, but intentionally does not account for global temporal shifts or warping. This design choice enables a focused investigation of reconstruction objectives without introducing alignment-related confounding factors.
Together, these considerations motivate the exploration of structure-aware reconstruction objectives as complementary alternatives to purely distance-based losses for time-series self-supervised representation learning.

\section{Related Works}

Recent work has questioned the effectiveness of transformers for time-series forecasting~\citep{zeng2023transformers}. This paper is related to the Dice Score Coefficient (DSC)~\citep{dice_score,dice_score2}, and the broader field of time-series modeling (TSM), specifically time-series pre-training models (TS-PTM).

\subsection{Evaluation Metrics}

Traditional TS-PTM research widely used reconstruction metrics such as MSE or MAE. These metrics ignore temporal misalignments, limiting their effectiveness. DTW~\citep{dtw} addressed misalignment but is computationally expensive. FastDTW~\citep{fastdtw} reduced complexity but is not differentiable and thus unsuitable for training. SoftDTW~\citep{softdtw} provided a differentiable approximation, making it usable in training. All remain distance-based. DILATE~\citep{DILATE} combines shape and temporal distortion losses, but is limited to forecasting.
Other alternatives target correlation. The Pearson Correlation Coefficient (PCC)~\citep{bishop2006pattern} measures linear dependence but is sensitive to phase shifts and mainly reflects point-wise similarity. In audio, Scale-Invariant SNR (SI-SNR)~\citep{luo2018tasnet} is used as a structure-aware objective, but it only maximizes signal-to-error ratio rather than comparing shapes directly.  
The Dice Score Coefficient (DSC)~\citep{dice_score,dice_score2} is widely used in semantic segmentation for overlap-based evaluation. We extend this idea to time-series and propose the SDSC. The SDSC directly measures structural overlap, emphasizing waveform shape rather than amplitude, and can serve as both a metric and a training loss. A summary of alignment-based objectives and our SDSC in terms of complexity and properties is provided in the Appendix~\ref{app:time_complexity}.

\subsection{Time-series Modeling}
According to~\citep{ssl_survey}, SSL typically follows a two-stage process: unsupervised pre-training followed by task-specific fine-tuning. This paradigm has driven major advances in time-series learning~\citep{signal_survey}.  

Several TS-PTM methods focus on representation learning. TS2Vec~\citep{ts2vec} learns contextual representations, while CoST~\citep{cost} applies contrastive objectives. TimesNet~\citep{timesnet} targets general time-series analysis, and~\citep{self_contrastive_representation} study semi-supervised settings. TI-MAE~\citep{ti-mae} introduces masked autoencoders, and SimMTM~\citep{simmtm} simplifies the masked framework. iTransformer~\citep{itransformer} uses inverted transformers for efficient forecasting, while the Unified Transformer~\citep{unifiedtransformer} supports multiple downstream tasks. TIMER~\citep{timer} develops generative pre-trained transformers. TimeSiam~\citep{timesiam} employs Siamese networks, and TimeDiT~\citep{timedit} explores diffusion-based pre-training. Cross-domain modeling has also been studied~\citep{large}, and~\citep{general} propose universal representations.  

Most of these advances rely on architectural design or contrastive strategies. Such approaches improve performance but are limited in capturing structural similarity. Unlike distance-based or alignment-based metrics, SDSC directly and efficiently quantifies structural similarity, addressing a crucial gap in representation learning for time-series. SimMTM~\citep{simmtm} is chosen as the baseline because it combines contrastive and reconstruction objectives in a modular manner.
In our setup, only the reconstruction loss (MSE) is replaced by SDSC, while the contrastive component (InfoNCE)~\citep{nce} remains fixed.
This controlled configuration ensures that observed performance differences originate from the reconstruction objective itself rather than from contrastive learning effects.

\section{Signal Dice Similarity Coefficient}
In this section, we introduce the SDSC, a new metric designed to explicitly quantify structural similarity between two signals. In signal representation learning, reconstructing signals accurately is important to capture their meaning.

\begin{figure}[ht]
    \centering
    \begin{subfigure}[b]{0.49\linewidth} 
        \centering
        \includegraphics[width=0.9\linewidth]{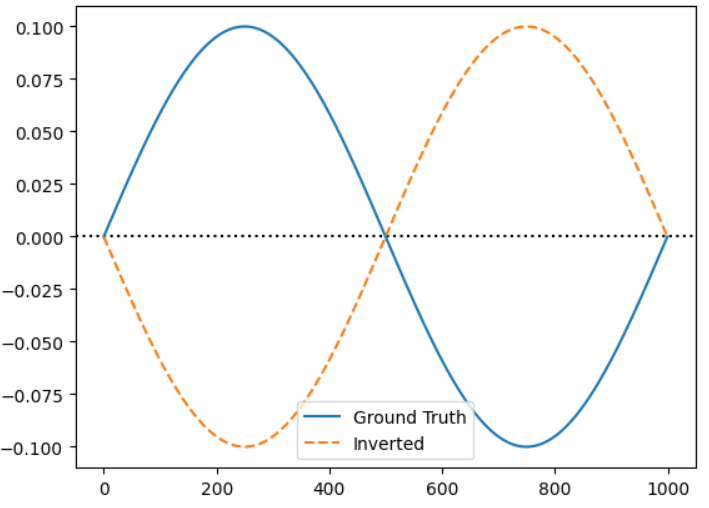} 
        \caption{Complete phase inversion under low-amplitude conditions.}
        \label{fig:motivation_a}
    \end{subfigure}
    \hfill 
    \begin{subfigure}[b]{0.49\linewidth} 
        \centering
        \includegraphics[width=0.9\linewidth]{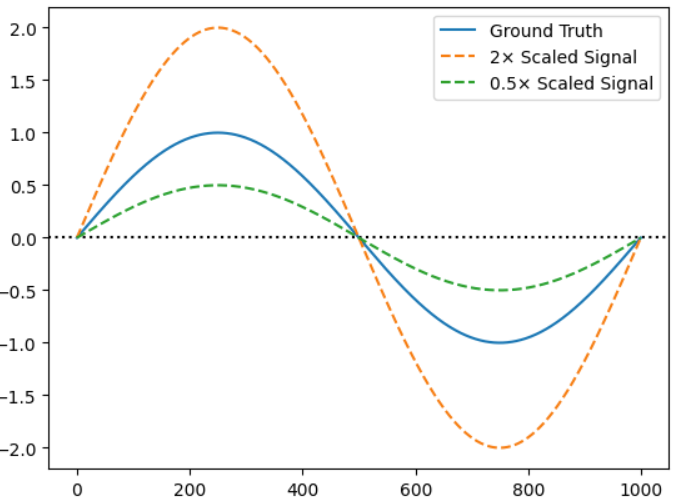}
        \caption{0.5$\times$ and 2$\times$ scaled signals introducing structural distortions.}
        \label{fig:motivation_b}
    \end{subfigure}
    
    \vspace{0.5em}
    \begin{subfigure}[b]{0.49\linewidth}
        \centering
        \includegraphics[width=0.9\linewidth]{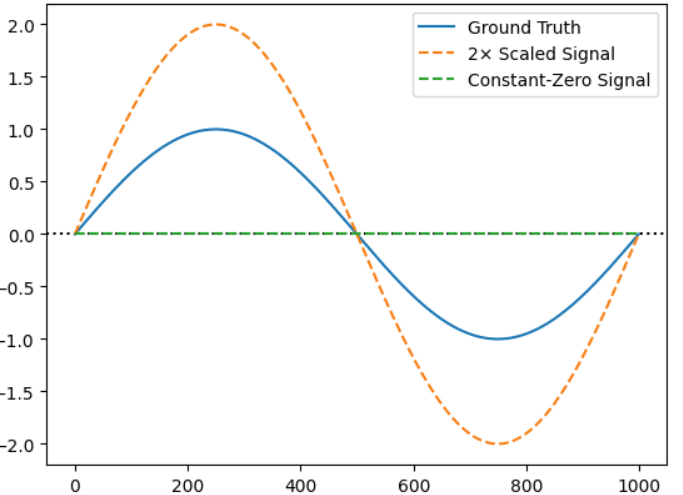}
        \caption{Constant zero signal vs. 2$\times$ scaled waveform with identical MSE.}
        \label{fig:motivation_c}
    \end{subfigure}
    \hfill
    \begin{subfigure}[b]{0.49\linewidth}
        \centering
        \includegraphics[width=0.9\linewidth]{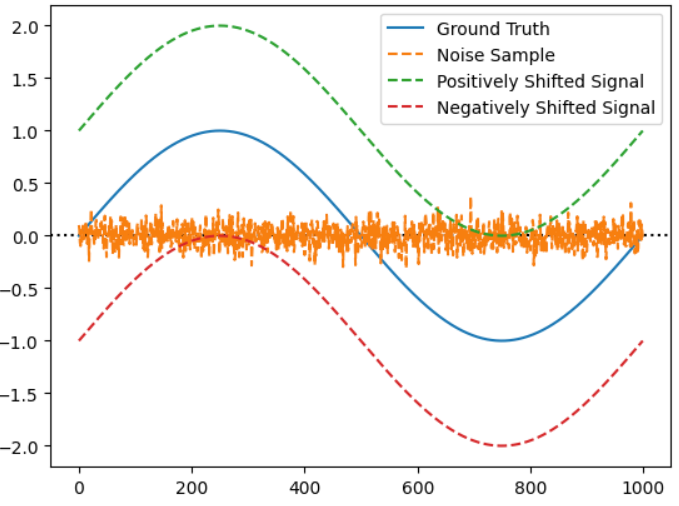}
        \caption{Noisy outputs with semantically valid yet shifted waveforms.}
        \label{fig:motivation_d}
    \end{subfigure}

    \caption{Examples demonstrating the limitations of distance-based metrics in capturing structural similarity. SDSC offers a more faithful assessment in (a) phase-shifted signals, (b) scale-induced distortions, (c) structurally dissimilar but MSE-equivalent signals, and (d) noisy outputs with underestimated errors.}
    \label{fig:motivation}
\end{figure}

\begin{table}[h!]
\caption{Quantitative evaluation of signal variations using distance-based metrics and the proposed SDSC.}
\label{table:motivation} 
\centering
\begin{tabular}{l S[table-format=1.4] S[table-format=1.2] S[table-format=1.2] S[table-format=1.2]}
\toprule
\textbf{Signals} & {\textbf{MSE↓}} & {\textbf{MAE↓}} & {\textbf{DTW↓}} & {\textbf{SDSC↑}} \\
\midrule
Inverted         & 0.0200 & 0.1272 & 0.0425 & 0.0000 \\
\addlinespace 
0.5x Scaled      & 0.1249 & 0.3180 & 0.1353 & 0.6667 \\
2x Scaled        & 0.4995 & 0.6360 & 0.2706 & 0.6667 \\
Zero             & 0.4995 & 0.6360 & 0.6360 & 0.0000 \\
\addlinespace
Noise Sample     & 0.5062 & 0.6361 & 0.2236 & 0.1137 \\
Positive Shifted & 1.0000 & 1.0000 & 0.6228 & 0.3887 \\
Negative Shifted & 1.0000 & 1.0000 & 0.6228 & 0.3887 \\
\bottomrule
\end{tabular}
\end{table}

\subsection{Distance-based Metrics}

Distance-based metrics such as MSE, mean absolute error (MAE) and dynamic time warping (DTW) are widely used to measure the difference between predicted and ground-truth signals. These metrics evaluate element-wise deviations and are effective in reducing numerical reconstruction errors. However, distance-based metrics focus primarily on signal amplitude and do not consider the polarity or structural shape of the waveform.

Figure~\ref{fig:motivation} and Table~\ref{table:motivation} illustrate examples that expose key limitations of distance-based metrics. Figure~\ref{fig:motivation_a} illustrates a complete phase inversion under low-amplitude conditions, visually preserving the waveform shape while reversing its polarity. As shown in Table~\ref{table:motivation} (Inverted), the inverted signal receives low error scores on all distance-based metrics (for example, MSE = 0.0200), making it appear as a high quality reconstruction despite its semantic reversal. Figure~\ref{fig:motivation_b} compares 0.5× and 2× scaled signals, both of which introduce comparable structural distortions but produce markedly different metric values due to amplitude differences. As indicated in Table~\ref{table:motivation}, the dependence on the amplitude obscures the true degree of structural deviation, resulting in an inaccurate evaluation of the signal quality. In Figure~\ref{fig:motivation_c}, a constant zero signal is evaluated alongside a 2× scaled waveform. As shown in Table~\ref{table:motivation}, both produce identical MSE scores (0.4995), despite their stark structural differences. The similarity in MSE scores despite structural differences reveals the inability of MSE to distinguish between waveforms when average magnitudes are equivalent. Lastly, Figure~\ref{fig:motivation_d} contrasts noisy outputs with semantically valid yet shifted waveforms. The noise-dominated signal produces an MSE (0.5062) comparable to semantically valid signals due to its fluctuation averaging around the baseline, making it appear deceptively accurate under distance-based metrics. Although numerically favorable, the output is structurally misaligned and functionally misleading. Such insensitivity to signal semantics is particularly problematic for physiological data like EEG or ECG, where subtle structural components often carry diagnostic significance. Therefore, exclusive reliance on amplitude-centric metrics may lead to semantically incorrect reconstructions.

\subsection{Definition of SDSC}

The SDSC extends the DSC, commonly used for set overlap in semantic segmentation, to continuous, signed time-series data. This extension is motivated by the observation that the goal of signal reconstruction in SSL is not merely to minimize the amplitude error, but to restore the signal's underlying shape, a concept that is difficult to formalize directly. We propose using the area under the curve as a tractable proxy for waveform shape. This reframes the problem of comparing two signal shapes as measuring the overlap between their respective areas. This area overlap problem is analogous to the well-posed problem of measuring pixel overlap in semantic segmentation, making the Dice Similarity Coefficient (DSC) a natural and theoretically sound foundation for our metric. Instead of relying on the membership of the sets, SDSC computes structural alignment from signed amplitude intersections at each time step. This formulation captures polarity agreement and local magnitude overlap, which represent the local structural consistency of the signals. It implicitly reflects small phase variations but does not account for temporal shifts or warping.  Like DSC, SDSC returns a score in the range $[0, 1]$. The original DSC measures set the similarity as follows:

\begin{equation}
DSC = \frac{2|X \cap Y|}{|X| + |Y|}
\end{equation}

Here, $|X|$ and $|Y|$ denote the cardinalities of the respective sets, and the metric reflects the size of their intersection relative to the total area. The SDSC extends this concept to the signal domain by interpreting the area under the curve as a proxy for the waveform structure. Given two signal functions $E(t)$ and $R(t)$ representing ground truth and reconstruction, the SDSC is defined as :

\begin{equation}
S(t) = E(t) \cdot R(t)
\label{eq:structure_same}
\end{equation}
\begin{equation}
\label{eq:check_minima}
M(t) = \frac{
\lvert E(t)\rvert + \lvert R(t)\rvert
- \bigl\lvert \lvert E(t)\rvert - \lvert R(t)\rvert \bigr\rvert
}{2}
\end{equation}
\begin{equation}
SDSC(E(t), R(t)) = \frac{ 2 \cdot \int  H(S(t)) \cdot M(t) \, dt }{ \int (|E(t)| + |R(t)|) \, dt }
\label{eq:original_sdsc}
\end{equation}

$H(\cdot)$ denotes the Heaviside step function(see in Appendix~\ref{app:heaviside}), and $t \in T$ is given time. The objective in signal representation learning is to maximize Equation (\ref{eq:original_sdsc}) toward 1. However, directly computing SDSC via integration is infeasible in practice, as real-world signals, such as EEG, lack known analytical expressions. To address this, a discrete approximation is adopted.

\begin{figure}[ht]
\centering
\includegraphics[width=\linewidth]{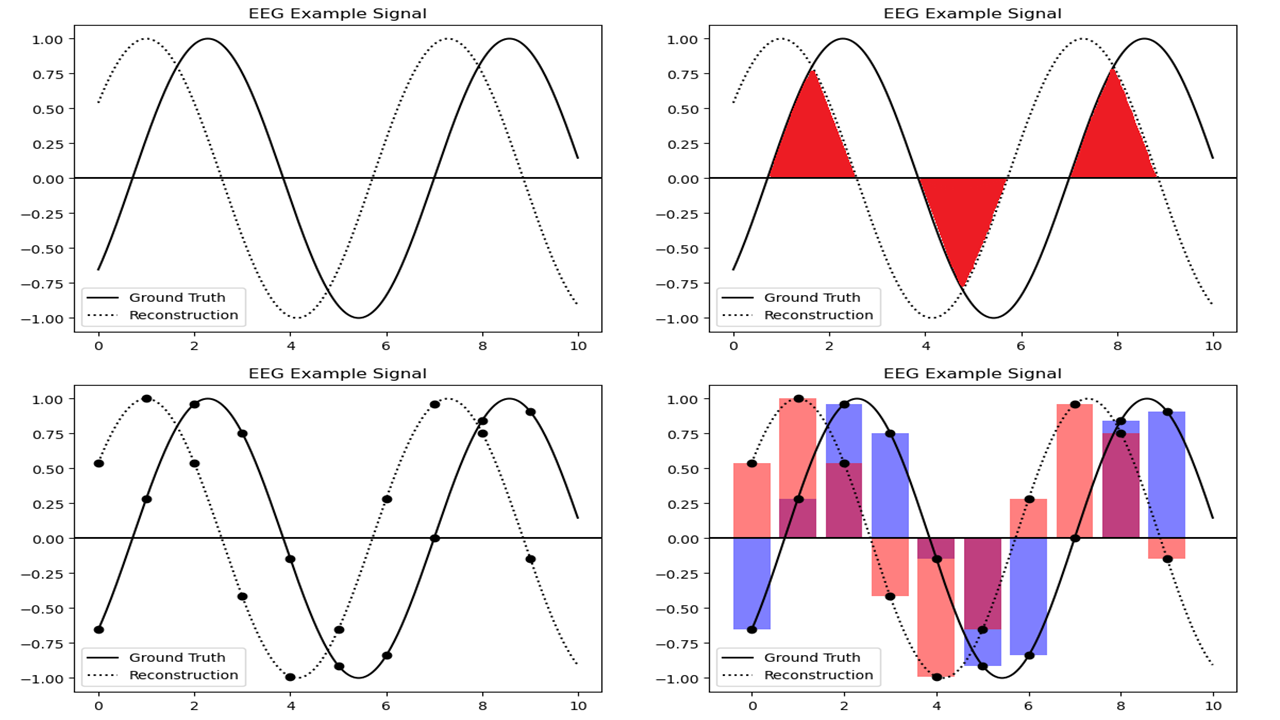}
\caption{ Example of intersection between two signals and discrete approximation.}
\label{fig:approx_sdsc}
\end{figure}

Figure \ref{fig:approx_sdsc} illustrates the approximation procedure. Although signals are continuous in nature, real-world signals are typically sampled at uniform intervals. Consequently, each sampled value is treated as a rectangle of unit width, allowing the continuous integral to be approximated by summation.

\begin{equation}
SDSC(E(t), R(t)) \approx \frac{  2 \cdot \sum (H(S(s)) \cdot M(s)) }{ \sum (|E(s)| + |R(s)|) +\epsilon}
\label{eq:approx_sdsc}
\end{equation}

where $s \in S$ are discrete sampling points, with $S \subset T$ and $\epsilon$ is a small constant to prevent division by zero. The proposed approximation enables a tractable computation of SDSC on real signals and ensures a consistent evaluation. As demonstrated in Table \ref{table:motivation}, the SDSC shows increased robustness to polarity shifts and amplitude scaling. As proven in Lemma~\ref{lem:bounded_sdsc}, the proposed SDSC is bounded in the range [0, 1] and facilitates standardized interpretation in all signal domains. Unlike MSE or DTW, the SDSC is less affected by signal magnitude, thereby reducing distortions due to scale and polarity. The normalized form also simplifies cross-domain comparisons, enabling a more structure-aware assessment of signal reconstruction quality.

\begin{table*}[ht]
\caption{Summary of pre-training performance averaged across forecasting and classification datasets. 
SDSC shows the most robust results overall (full results in Appendix~\ref{app:full_pre_forecasting_results},~\ref{app:full_pre_classification_results}). 
Note: SI-SNR values use a different scale and sometimes fail to converge (e.g., ETTh1), so they are reported for completeness.} 
\label{tab:summary_results}
\centering
\small 
\begin{tabular}{ll S[table-format=3.4] S[table-format=1.4] S[table-format=1.4]}
\toprule
\textbf{Dataset} & \textbf{Loss} & {MSE↓} & {MAE↓} & {SDSC↑} \\
\midrule
\multirow{5}{*}{\textbf{Avg (Forecasting)}} 
 & MSE    & 0.4852 & 0.3525  & 0.7670\\
 & SoftDTW & 1.3273 & 0.7432  & 0.4990 \\
 & PCC    & 1.3289 & 0.6705 & 0.5274 \\
 & SI-SNR & 34.9085& 2.5408 & 0.4523 \\
\cmidrule(lr){2-5}
 & \textbf{SDSC(Ours)}   & 0.6348 & 0.3870  & 0.7723\\
 & \textbf{Hybrid(Ours)} & \textbf{0.4783} & \textbf{0.3368} & \textbf{0.7841} \\
\midrule
\multirow{5}{*}{\textbf{Avg (Classification)}} 
 & MSE    & 50.3203 & 3.5269 & 0.6105 \\
 & Soft-DTW & \textbf{49.1339} & \textbf{3.4751} & 0.6210  \\
 & PCC   & 120.0105 & 4.5091 & 0.1622 \\
 & SI-SNR & 118.6110 & 4.4846 & 0.1693 \\
\cmidrule(lr){2-5}
 & \textbf{SDSC(Ours)}   & 74.0253 & 3.8626 & \textbf{0.6610}\\
 & \textbf{Hybrid(Ours)} & 50.3471 & 3.5286 & 0.6481 \\
\bottomrule
\end{tabular}
\end{table*}

\subsection{Hybrid Loss Integration}

Since the SDSC score is bounded in $[0,1]$, we can define the loss as $1 - SDSC(\cdot)$.

\begin{equation}
    \mathcal{L}_{sdsc} = 1 - SDSC(E(S), R(S))
\label{eq:SDSC_Loss}
\end{equation}

However, the use of the Heaviside step function in Equation (\ref{eq:original_sdsc}) introduces discontinuities, which can negatively affect the stability of training. Continuity is preserved when at least one of the signals maintains the same sign at the corresponding sampled points. However, the likelihood of sign mismatches increases when the sampling resolution is low. To enable stable gradient-based optimization, a smooth approximation of the Heaviside function is introduced. The following sigmoid-based formulation is used, with a sharpness parameter $\alpha$.

\begin{equation}
\hat{H}(x) = \frac{1}{1 + e^{-\alpha x}} \quad
\label{eq:sigmoid_heaviside}
\end{equation}

If the sharpness parameter $\alpha$ is large, the sigmoid-based approximation $\hat{H}(x)$ more closely resembles the original Heaviside function. However, excessively large values of $\alpha$ can lead to sharp transitions that result in unstable gradients, potentially damaging the training process.

SDSC captures structure but ignores amplitude, whereas MSE captures amplitude but misses structure. To balance the strengths of both approaches, this work proposes a hybrid loss function that combines the structural awareness of SDSC with the amplitude sensitivity of MSE. The final objective function is formulated as follows:

\begin{equation}
\mathcal{L}_{hybrid} = \lambda_{sdsc} \cdot \mathcal{L}_{sdsc} + \lambda_{mse} \cdot \mathcal{L}_{MSE}
\label{eq:hybrid_loss}
\end{equation}

Here, $\lambda_{sdsc}$ and $\lambda_{mse}$ are parameters that control the trade-off between structural accuracy and amplitude-based accuracy. To determine these weights, we adopt the uncertainty-based tuning strategy proposed in ~\citep{awl}, where the weighting coefficients are adapted based on the homoscedastic uncertainty associated with each loss term. This hybrid formulation promotes reconstructions that are structurally aligned and numerically precise.In practice, each loss term’s weight is parameterized by a trainable log-variance term following \citet{awl}, and updated jointly with model parameters to balance the relative homoscedastic uncertainty of $\mathcal{L}_{sdsc}$ and $\mathcal{L}_{mse}$.

The overall pre-training objective of SimMTM consists of a contrastive term $\mathcal{L}_{con}$ (InfoNCE)~\citep{nce} and a reconstruction term $\mathcal{L}_{rec}$.
We keep $\mathcal{L}_{con}$ identical to the original SimMTM formulation and replace $\mathcal{L}_{rec}$ with either MSE, SDSC, or the proposed Hybrid loss.
The total loss is given by:
\begin{equation}
\mathcal{L}_{total} = \mathcal{L}_{con} + \mathcal{L}_{rec},
\label{eq:total_loss}
\end{equation}
where $\mathcal{L}_{rec} \in \{\mathcal{L}_{MSE}, \mathcal{L}_{sdsc}, \mathcal{L}_{hybrid}, \mathcal{L}_{pcc}, \mathcal{L}_{si\_snr},\mathcal{L}_{softdtw}\}$. This formulation isolates the effect of the reconstruction objective under a fixed contrastive setup.

\section{Experiments}

All experiments are conducted with fixed random seeds to ensure reproducibility. The contrastive objective of SimMTM is kept identical across all experiments; therefore, any observed differences in downstream performance can be attributed solely to the reconstruction objective (MSE, SDSC, or Hybrid), rather than to changes in the contrastive component. All training and evaluation are performed on two NVIDIA RTX 3090 GPUs.

For a controlled comparison, we adopt SimMTM~\citep{simmtm} as the backbone model throughout all experiments. SimMTM is conceptually lightweight in its framework design, but internally employs transformer-based encoders with multi-head self-attention and temporal masking, making it comparable in expressive capacity to recent transformer-based pretraining models such as PatchTST~\citep{patchtst}. Notably, SimMTM was reported in the NeurIPS 2023 benchmark suite to achieve competitive or superior performance compared to several transformer backbones, including PatchTST, indicating that its simplicity does not imply limited representational power. Using a single backbone allows us to isolate the effect of the reconstruction objective without confounding architectural factors.

In our setup, SDSC replaces the MSE reconstruction loss, while the contrastive objective remains unchanged. This design enables a clean analysis of how different reconstruction objectives influence representation learning under identical contrastive conditions. In addition to MSE, we include several structure-aware baselines for comparison, including Pearson correlation coefficient (PCC)~\citep{bishop2006pattern} and scale-invariant signal-to-noise ratio (SI-SNR)~\citep{luo2018tasnet}. All baseline models are reproduced using their official implementations. Detailed hyperparameter settings for all experiments, including the choice of $\alpha = 10$ for SDSC based on the sensitivity analysis in Appendix~\ref{app:gradient_sensitivity}, are provided in Appendix~\ref{app:hyperparameters}.

For forecasting tasks, mean squared error (MSE) and mean absolute error (MAE) are used as evaluation metrics. For classification tasks, accuracy, precision, recall, and F1 score are reported along with their macro-averaged values. All time-series inputs are z-score normalized per channel using statistics computed exclusively from the training split, ensuring scale consistency and removing DC offsets without introducing data leakage.

\subsection{Pre-training}

During the pre-training stage, we compare three reconstruction objectives: MSE, SDSC, and a hybrid objective that combines both. Table~\ref{tab:summary_results} summarizes the pre-training results across forecasting and classification datasets.

As expected, models trained with MSE achieve lower reconstruction error under distance-based metrics, while SDSC-based models attain higher SDSC scores, reflecting stronger structural agreement. The two metrics exhibit only weak correlation, suggesting that they capture complementary aspects of the signal. This motivates a closer examination of how structural alignment and amplitude fidelity interact during pre-training and how these properties translate to downstream performance.

\begin{figure}[ht]
    \centering
    
    \begin{subfigure}[b]{\linewidth} 
        \centering
        \includegraphics[width=0.5\linewidth]{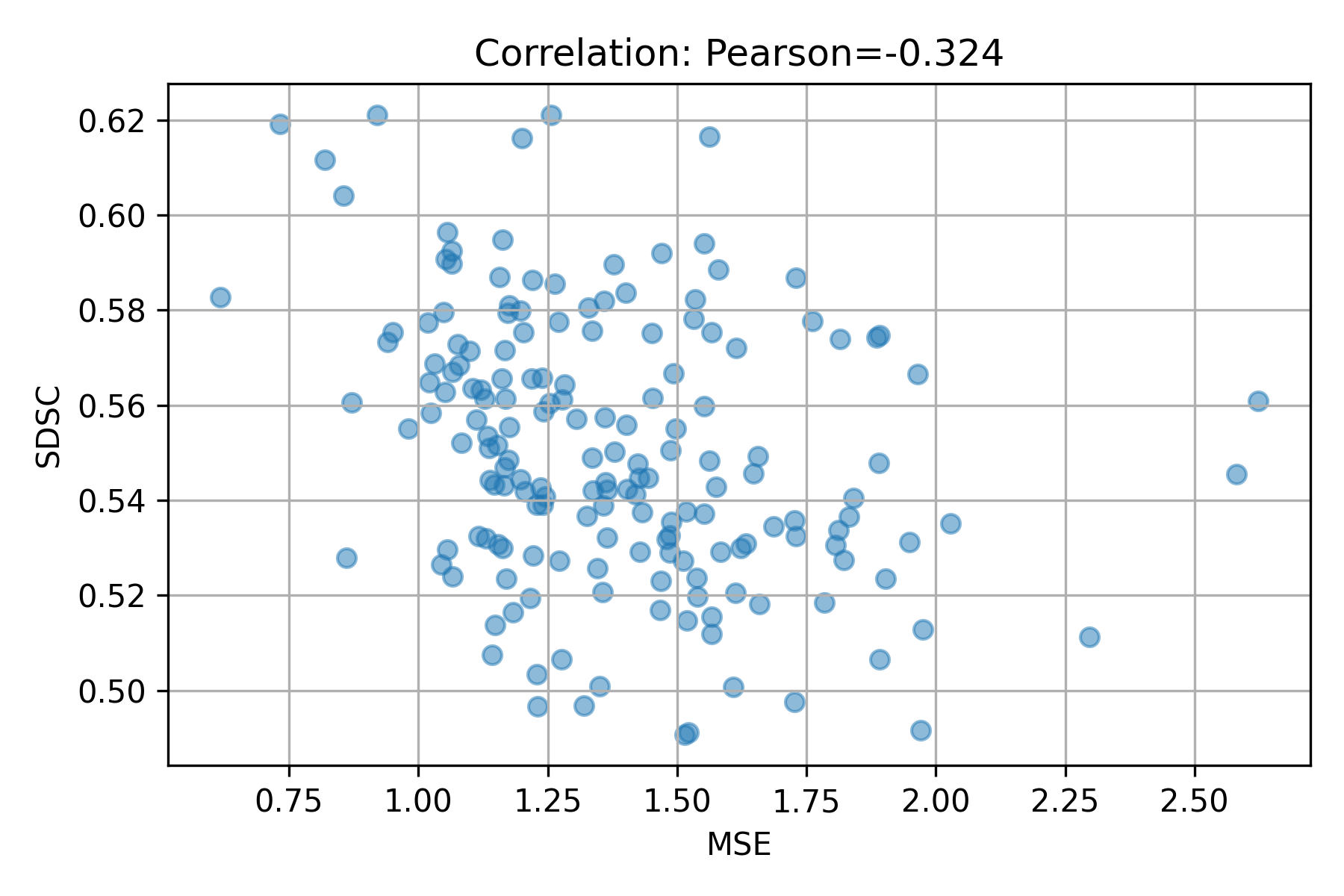} 
        \caption{Scatter plot between MSE and SDSC under MSE-based pre-training.}
        \label{fig:pre-training_corr}
    \end{subfigure}

    \vspace{0.5em} 

    \begin{subfigure}[b]{0.49\linewidth}
        \centering
        \includegraphics[width=0.7\linewidth]{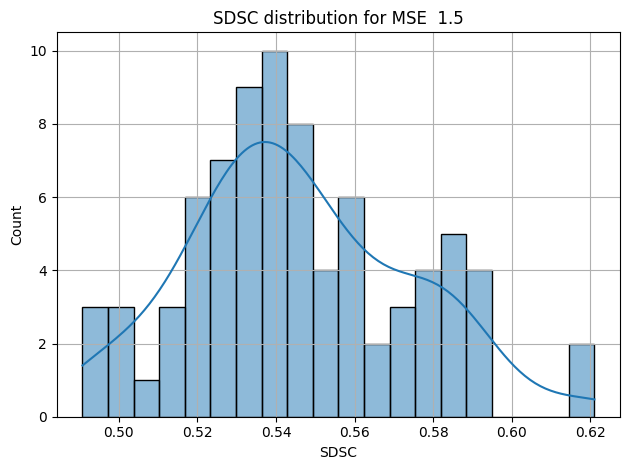}
        \caption{MSE-based model.}
        \label{fig:dist_mse}
    \end{subfigure}
    \hfill 
    \begin{subfigure}[b]{0.49\linewidth}
        \centering
        \includegraphics[width=0.7\linewidth]{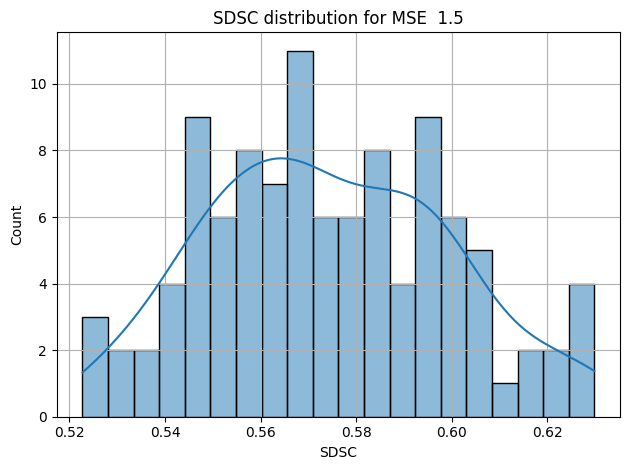}
        \caption{SDSC-based model.}
        \label{fig:dist_sdsc}
    \end{subfigure}

    \caption{Analysis of structural alignment in MSE-based pre-training. (a) The weak negative correlation between MSE and SDSC suggests limited alignment. (b, c) At a fixed MSE of $1.5\pm\epsilon$, the SDSC-based model achieves a distribution with higher structural scores.}
    \label{fig:correlation_and_distribution}
\end{figure}

Figure~\ref{fig:pre-training_corr} shows the relationship between MSE and SDSC for models pre-trained on ETTh1 using MSE. SDSC increases as MSE decreases, but the Pearson correlation is $-0.324$, indicating a weak alignment. Figure~\ref{fig:dist_mse}, Figure~\ref{fig:dist_sdsc}  compare the SDSC distributions at fixed MSE ($1.5\pm\epsilon$) under MSE-based and SDSC-based pre-training. SDSC-based pre-training achieves higher SDSC values at the same MSE level. 

\begin{table}[h]
\caption{Comparison of SDSC concentration under a fixed MSE ($1.5\pm\epsilon$).}
\label{tab:sdsc_concentration}
\centering
\begin{tabular}{l S[table-format=1.4] S[table-format=1.4]}
\toprule
\textbf{Model Type} & {\textbf{Std Dev}} & {\textbf{IQR}} \\
\midrule
MSE-based model    & 0.0280 & 0.0418 \\
SDSC-based model   & 0.0249 & 0.0384 \\
\bottomrule
\end{tabular}
\end{table}

Figure~\ref{fig:pre-training_corr} illustrates the relationship between MSE and SDSC for models pre-trained on ETTh1 using MSE. Although SDSC generally increases as MSE decreases, the Pearson correlation coefficient is $-0.324$, indicating limited alignment between the two objectives. Figures~\ref{fig:dist_mse} and~\ref{fig:dist_sdsc} compare the distributions of SDSC scores under a fixed MSE constraint ($1.5\pm\epsilon$) for MSE-based and SDSC-based pre-training, respectively. Under identical MSE conditions, SDSC-based pre-training consistently yields higher SDSC values.

Table~\ref{tab:sdsc_concentration} further reports the standard deviation and interquartile range (IQR) of SDSC scores under fixed MSE. SDSC-based models exhibit tighter distributions, indicating more consistent structural alignment across samples. These results suggest that while MSE-based pre-training captures certain structural features implicitly, it does so unreliably. In contrast, SDSC explicitly enforces structural consistency at the cost of increased reconstruction error. The hybrid objective provides a balanced trade-off, maintaining stable behavior across both structural and amplitude-based metrics.

\subsection{Forecasting}

\begin{table}[H]
\caption{Forecasting fine-tuning performance on a representative dataset (Electricity) and the average across all datasets. Our method shows strong performance on complex datasets and achieves the best overall average. (Full results  in Appendix~\ref{app:full_forecasting_results}.)}
\label{tab:forecasting_summary}
\centering
\small 
\begin{tabular}{ll S[table-format=1.3] S[table-format=1.3]}
\toprule
\textbf{Dataset} & \textbf{Pre-training Loss} & {\textbf{MSE↓}} & {\textbf{MAE↓}} \\
\midrule
\multirow{5}{*}{Electricity}
 & MSE    & 0.200 & 0.291 \\
 & Soft-DTW & 0.200 & 0.290 \\
 & PCC    & 0.199 & \textbf{0.288} \\
 & SI-SNR & 0.203 & 0.292 \\
\cmidrule(lr){2-4}
 & \textbf{SDSC(Ours)}   & 0.200 & 0.293 \\
 & \textbf{Hybrid(Ours)} & \textbf{0.198} & \textbf{0.288} \\
\midrule
\multirow{5}{*}{Avg}
 & MSE    & 0.295 & \textbf{0.316} \\
 & Soft-DTW & 0.303 & 0.322 \\
 & PCC    & 0.296 & 0.318 \\
 & SI-SNR & 0.310 & 0.325 \\
\cmidrule(lr){2-4}
 & \textbf{SDSC(Ours)}   & \textbf{0.294} & \textbf{0.316} \\
 & \textbf{Hybrid(Ours)} & \textbf{0.294} & \textbf{0.316} \\
\bottomrule
\end{tabular}
\end{table}

\begin{table*}[ht!]
\caption{Summary of freeze classification performance, averaged across scenarios. Our proposed SDSC shows notable improvements in the in-domain setting. (full results in Appendix~\ref{app:full_freeze_classification_results})}
\label{tab:freeze_classification_summary}
\centering
\small
\sisetup{detect-weight=true, detect-family=true}

\begin{tabular}{l l S[table-format=2.2] S[table-format=2.2] S[table-format=2.2] S[table-format=2.2] S[table-format=2.2]}
\toprule
\textbf{Scenario} & \textbf{Loss} & {\textbf{Acc.↑}} & {\textbf{Prec.↑}} & {\textbf{Rec.↑}} & {\textbf{F1↑}} & {\textbf{Avg↑}} \\
\midrule
\multirow{5}{*}{\textbf{Avg (In Domain)}}
 & MSE    & 75.45 & 73.07 & 63.49 & 64.59 & 69.15 \\ 
 & Soft-DTW & 68.76 & 40.89 & 47.36 & 43.52 & 50.13 \\
 & PCC    & 71.40 & 47.38 & 51.56 & 46.70 & 54.26 \\
 & SI-SNR & 71.08 & 47.40 & 51.81 & 46.91 & 54.30 \\
\cmidrule(lr){2-7}
 & \textbf{SDSC(Ours)}   & \textbf{76.38} & \textbf{74.19} & \textbf{64.93} & \textbf{65.85} & \textbf{70.34} \\
 & \textbf{Hybrid(Ours)} & 76.23 & 72.72 & 65.61 & 66.47 & 70.26 \\
\midrule
\multirow{5}{*}{\textbf{Avg (Cross Domain)}}
 & MSE    & \textbf{62.19} & 39.04 & \textbf{47.89} & 41.42 & 47.63 \\ 
 & Soft-DTW & 62.00 & \textbf{41.47} & 47.75 & \textbf{41.73} & \textbf{48.24} \\
 & PCC    & 60.47 & 39.22 & 46.18 & 39.19 & 46.27 \\
 & SI-SNR & 60.13 & 38.55 & 45.99 & 38.35 & 45.75 \\
\cmidrule(lr){2-7}
 & \textbf{SDSC(Ours)}   & 61.64 & 39.75 & 47.38 & 40.34 & 47.28 \\
 & \textbf{Hybrid(Ours)} & 62.00 & 39.06 & 47.75 & 41.45 & 47.70 \\
\bottomrule
\end{tabular}
\end{table*}

\begin{table*}[ht!]
\caption{Summary of fine-tuning classification performance, averaged across scenarios. (full results in Appendix~\ref{app:full_classification_results}.)}
\label{tab:finetuning_classification_summary}
\centering
\small
\sisetup{detect-weight=true, detect-family=true}
\begin{tabular}{l l S[table-format=2.2] S[table-format=2.2] S[table-format=2.2] S[table-format=2.2] S[table-format=2.2]}
\toprule
\textbf{Scenario} & \textbf{Loss} & {\textbf{Acc.↑}} & {\textbf{Prec.↑}} & {\textbf{Rec.↑}} & {\textbf{F1↑}} & {\textbf{Avg↑}} \\
\midrule
\multirow{5}{*}{\textbf{Avg (In Domain)}}
 & MSE    & 79.66 & 75.87 & 71.24 & 71.09 & 74.46 \\ 
 & Soft-DTW & 79.07 & 75.68 & 70.01 & 69.43 & 73.55 \\
 & PCC    & \textbf{79.76} & \textbf{76.17} & \textbf{71.34} & \textbf{71.21} & \textbf{74.62} \\
 & SI-SNR & 79.31 & 75.85 & 70.27 & 70.31 & 73.94 \\
\cmidrule(lr){2-7}
 & \textbf{SDSC(Ours)}   & 79.60 & 76.01 & 70.54 & 70.69 & 74.21 \\
 & \textbf{Hybrid(Ours)} & 79.52 & 75.60 & 71.10 & 70.21 & 74.11 \\
\midrule
\multirow{5}{*}{\textbf{Avg (Cross Domain)}}
 & MSE    & 83.74 & 85.46 & \textbf{84.89} & \textbf{84.33} & \textbf{84.65} \\ 
 & Soft-DTW & 83.20 & 84.40 & 84.55 & 83.89 & 84.05 \\
 & PCC    & 83.79 & 86.78 & 82.37 & 83.03 & 83.99 \\
 & SI-SNR & \textbf{84.27} & 84.75 & 84.13 & 83.19 & 84.09 \\
\cmidrule(lr){2-7}
 & \textbf{SDSC(Ours)}   & 83.27 & \textbf{85.93} & 81.33 & 82.62 & 83.29 \\
 & \textbf{Hybrid(Ours)} & 83.23 & 85.09 & 83.35 & 82.98 & 83.66 \\
\bottomrule
\end{tabular}
\end{table*}

Forecasting experiments are conducted in an in-domain setting using models pre-trained with different reconstruction objectives. Evaluation is performed at the best checkpoint for both pre-training and fine-tuning. Following standard practice, MSE and MAE are used as evaluation metrics.

Table~\ref{tab:forecasting_summary} summarizes the forecasting results on the Electricity dataset and the average across all datasets. As shown in Appendix~\ref{app:full_pre_forecasting_results}, models pre-trained with MSE implicitly learn a degree of structural alignment, while SDSC-based models, despite exhibiting higher reconstruction error, achieve downstream forecasting performance comparable to MSE-based models. This observation suggests that beyond a certain point, further reduction in reconstruction error yields diminishing returns for forecasting accuracy.

Notably, SDSC-based models achieve similar predictive performance despite substantially higher MSE, indicating that preserving local waveform structure alone can be sufficient for downstream forecasting. The hybrid objective consistently maintains competitive performance, reflecting its ability to balance amplitude fidelity and structural consistency across datasets.

\subsection{Classification}

Classification experiments are conducted in both in-domain and cross-domain settings to assess the generalization behavior of the learned representations. We evaluate two scenarios: frozen encoders, which isolate the effect of pre-training, and end-to-end fine-tuning, where pre-training serves as initialization. All models are evaluated using accuracy, precision, recall, and F1 score.

Table~\ref{tab:freeze_classification_summary} and Appendix~\ref{app:full_freeze_classification_results} report the results with frozen encoders, while Table~\ref{tab:finetuning_classification_summary} and Appendix~\ref{app:full_classification_results} summarize the fine-tuning results. When encoders are frozen, SDSC-based pre-training consistently achieves stronger performance in in-domain settings, where preserving structural fidelity is particularly important. In cross-domain scenarios, performance differences vary across datasets, reflecting differences in signal characteristics. For example, datasets dominated by amplitude-dependent features favor MSE-based pre-training, whereas datasets emphasizing waveform structure benefit more from SDSC-based objectives.

With end-to-end fine-tuning, performance gaps between objectives narrow substantially, indicating that fine-tuning can override pre-training differences. Nevertheless, SDSC-based models consistently achieve higher precision across multiple settings, suggesting that enforcing structural consistency during pre-training leads to representations that are more selective and semantically coherent. The hybrid objective provides a robust compromise across diverse datasets, offering balanced performance without requiring task-specific tuning.

\section{Conclusions}

In this paper, we introduced the Signal Dice Similarity Coefficient (SDSC), a structure-aware reconstruction metric for semantic representation learning in time-series data. In this work, \emph{structure-aware} specifically refers to local waveform consistency characterized by pointwise sign agreement and magnitude overlap, rather than global alignment or phase warping. Unlike conventional distance-based metrics, SDSC is bounded within the normalized range $[0,1]$ and exhibits reduced sensitivity to amplitude variation. Since SDSC is not directly differentiable, we proposed a smooth approximation that enables its use as a training objective. We further identified scenarios in which purely structure-focused objectives may overlook amplitude-critical signal properties, and addressed this limitation through a hybrid loss that combines SDSC with MSE.

Our experimental results show that SDSC enhances local structural fidelity in the reconstruction branch under identical contrastive settings, thereby contributing complementary information to self-supervised representation learning. While the observed improvements are generally moderate, they consistently indicate that structure-aware reconstruction influences representation quality in a manner that is not fully captured by distance-based losses alone. Importantly, comparable downstream performance between MSE- and SDSC-based models should not be interpreted as evidence of MSE superiority. Rather, it suggests that amplitude-based metrics may overestimate reconstruction quality by assigning low errors to semantically inconsistent signals. SDSC exposes such cases more explicitly, revealing limitations of purely distance-based objectives and highlighting the compatibility of structural and amplitude-based criteria.

Across forecasting tasks, SDSC-based models often incur higher reconstruction error in terms of MSE, yet achieve comparable downstream performance under identical contrastive conditions. This observation suggests that excessive minimization of reconstruction error yields diminishing returns, and that enforcing local structural alignment may be sufficient for learning effective representations. In classification tasks, SDSC consistently improves in-domain performance when encoders are frozen, indicating that structure-aware pre-training better preserves semantic information relevant to downstream discrimination. While alignment-based objectives such as SoftDTW or DILATE can offer advantages in certain forecasting settings, their quadratic computational complexity limits practical scalability. In contrast, our complexity analysis highlights SDSC as a lightweight, alignment-free alternative that achieves competitive downstream performance with substantially lower computational cost.

Overall, our findings question the default reliance on MSE as a reconstruction objective for time-series self-supervised learning and position SDSC as a practical metric for structure-aware representation learning. Future work may explore integrating SDSC into other self-supervised frameworks and investigating its role in domain adaptation or cross-modal learning scenarios. Understanding the relationship between structural similarity and task-specific generalization also remains an open research question. We leave direct head-to-head training with alignment-based objectives such as DILATE, as well as integration into additional pretraining frameworks (e.g., TiMAE or alternative contrastive formulations), as future work due to computational constraints. Finally, we provide a practical guideline in Appendix~\ref{app:pratical_guidline} summarizing when SDSC, MSE, or the hybrid objective is most appropriate under different learning regimes.

\nocite{langley00}

\bibliography{example_paper}
\bibliographystyle{icml2026}

\newpage
\appendix

\onecolumn
\clearpage

\section{Appendix}

\subsection{Reproducibility Statement}
All source code for our experiments is available in an anonymous GitHub repository to
ensure full reproducibility. This repository includes the implementation of our proposed
SDSC loss function and the scripts to reproduce all forecasting and classification results presented in Section 4. The key hyperparameters for all pre-training and fine-tuning experiments are detailed in Appendix ~\ref{app:hyperparameters}. The mathematical derivation of SDSC is provided in
Section 3.2, and a formal proof of its boundedness is available in Appendix ~\ref{lem:bounded_sdsc}. https://anonymous.4open.science/r/SignalDiceSimilarityCoefficient-75A4

\subsection{Computational Complexity Analysis}
\label{app:time_complexity}

Alignment-based objectives such as DTW, FastDTW, SoftDTW, and DILATE have been widely used to handle temporal misalignments in time-series. 
However, these approaches are fundamentally distance-based and typically quadratic in sequence length $T$, which can be prohibitive for large-scale pre-training. 
FastDTW reduces the complexity to linear time but remains approximate and non-differentiable, limiting its applicability for gradient-based optimization. 
SoftDTW provides a differentiable relaxation but still incurs $O(T^2)$ complexity. 
DILATE combines shape and temporal distortion terms, yet inherits the same quadratic cost and has been applied only in forecasting tasks.

\begin{table}[h]
\centering
\caption{Comparison of structure-aware metrics for time-series. $T$ denotes sequence length. SDSC differs by being alignment-free, lightweight, and interpretable.}
\label{table:complexity}
\renewcommand{\arraystretch}{1.2} 
\begin{tabular}{@{}lccl@{}}
\toprule
\textbf{Method} & \textbf{Type} & \textbf{Time Complexity} & \textbf{Remarks} \\
\midrule
PCC & Correlation-based & $O(T)$ & \makecell[l]{Measures linear dependence, sensitive \\to phase shift} \\
SI-SNR & Ratio-based & $O(T)$ & \makecell[l]{Structure-aware in audio, indirect via \\signal-to-error ratio} \\
DTW & Distance-based & $O(T^2)$ & Exact alignment, non-differentiable \\
FastDTW & Distance-based & $O(T) \text{ (approx.)}$ & Efficient but approximate, non-differentiable \\
SoftDTW & Distance-based & $O(T^2)$ & Differentiable relaxation of DTW \\
DILATE & Distance-based & $O(T^2)$ & \makecell[l]{Combines shape and temporal distortion, \\ tailored for forecasting} \\
\midrule
\textbf{SDSC (ours)} & Similarity-based & $O(T)$ & \makecell[l]{Overlap-based, bounded in [0,1], \\ differentiable via sigmoid} \\
\bottomrule
\end{tabular}
\end{table}

In contrast, our proposed SDSC is an alignment-free similarity measure. 
It operates in linear time $O(T)$, is bounded in $[0,1]$, and can be smoothly approximated for differentiation, making it lightweight and practical for representation learning at scale. 
Table~\ref{table:complexity} summarizes the key properties of these methods.

\subsection{Heaviside convention}
\label{app:heaviside}

\begin{equation}
H(x) =
\begin{cases}
1, & x > 0, \\
0, & x \leq 0.
\end{cases}
\end{equation}

This convention sets $H(0)=0$, which we adopt as the default throughout both evaluation and training.
During optimization, we use a smooth sigmoid relaxation $\hat{H}(x) = \frac{1}{1+e^{-\alpha x}}$ with $\alpha=10$, so that $\hat{H}(x) \approx H(x)$ while remaining differentiable.

\clearpage
\subsection{Gradient Sensitivity Analysis}

\label{app:gradient_sensitivity} 

In this section, we measure the gradient norm with respect to different types of signal perturbations to assess how each loss function responds to structural variations. The analysis is based on the same toy cases as in Table~\ref{table:motivation}, including the jitter case. 

\begin{table}[ht!]
\caption{Analysis of gradient sensitivity. (a) Comparison across different loss functions. (b) Effect of the sharpness parameter $\alpha$ for the SDSC loss.}
\label{tab:gradient_sensitivity_combined}
\centering

\begin{subtable}{0.48\linewidth}
    \centering
    \caption{Gradient Sensitivity}
    \label{tab:sensitivity}
    \begin{tabular}{l S[table-format=1.4] S[table-format=1.4] S[table-format=1.4]}
    \toprule
    \textbf{Example} & {\textbf{MSE}} & {\textbf{MAE}} & {\textbf{SDSC}} \\
    \midrule
    Inverted     & 0.0894 & 0.0316 & 0.0000 \\
    \addlinespace
    0.5x Scaled  & 0.0223 & 0.0316 & 0.0442 \\
    2x Scaled    & 0.0447 & 0.0316 & 0.0110 \\
    Zero         & 0.0447 & 0.0316 & 0.0000 \\
    \addlinespace
    Noise Sample & 0.0194 & 0.0316 & 0.0237 \\
    Shifted      & 0.0632 & 0.0316 & 0.0075 \\
    \addlinespace
    Jittered     & 0.0032 & 0.0316 & 0.0248 \\
    \bottomrule
    \end{tabular}
\end{subtable}
\hfill 
\begin{subtable}{0.48\linewidth}
    \centering
    \caption{Sensitivity to $\alpha$}
    \label{tab:alpha_sensitivity}
    \begin{tabular}{l S[table-format=1.4] S[table-format=1.4] S[table-format=1.4]}
    \toprule
    \textbf{Example} & {\textbf{$\alpha=1$}} & {\textbf{$\alpha=10$}} & {\textbf{$\alpha=100$}} \\
    \midrule
    Inverted     & 0.0091 & 0.0082 & 0.0047 \\
    \addlinespace
    0.5x Scaled  & 0.0289 & 0.0437 & 0.0436 \\
    2x Scaled    & 0.0062 & 0.0102 & 0.0102 \\
    Zero         & 0.0000 & 0.0000 & 0.0000 \\
    \addlinespace
    Noise Sample & 0.0152 & 0.0228 & 0.0237 \\
    Shifted      & 0.0074 & 0.0087 & 0.0076 \\
    \addlinespace
    Jittered     & 0.0165 & 0.0228 & 0.0242 \\
    \bottomrule
    \end{tabular}
\end{subtable}

\end{table}

Tables~\ref{tab:sensitivity} and~\ref{tab:alpha_sensitivity} present the gradient norms for MSE, MAE, and SDSC, and the effect of varying the $\alpha$ parameter in SDSC, respectively.

MSE exhibits significant gradient changes under amplitude perturbations, but it fails to respond meaningfully to minor structural variations, such as jitter. In contrast, SDSC yields low gradients for structure-preserving signals (e.g., shifted), while responding with larger gradients when local waveform patterns are distorted (e.g., jittered). Although gradient vanishing may occur in some structural-breaking cases, SDSC remains robust to amplitude shifts.

As $\alpha$ increases, the behavior of the approximated SDSC becomes more similar to that of the original formulation. The results suggest that $\alpha=10$ is sufficient for a close approximation in practice.

\subsection{Downstream Task Analysis with \texorpdfstring{$\alpha$}{alpha}}

\begin{table}[h]
\centering
\caption{Impact of the sharpness parameter $\alpha$ on pre-training reconstruction performance across
scenarios}
\label{table:impact_of_alpha}
\renewcommand{\arraystretch}{1.2} 
\begin{tabular}{@{}llccc@{}}
\toprule
\textbf{Scenario} & \textbf{Example} & {\textbf{MSE↓}} & {\textbf{MAE↓}} & \textbf{SDSC↑} \\
\midrule
\multirow{3}{*}{Forecasting} 
& $\alpha=1$ & 1.7030 & 0.9112 & 0.3946 \\
& $\alpha=10$ & 1.7230 & 0.9140 & 0.3975 \\
& $\alpha=100$ & 1.7410 & 0.9177 & 0.3995 \\
\midrule
\multirow{3}{*}{Classification} 
& $\alpha=1$ & 0.1012 & 0.2406 & 0.6808 \\
& $\alpha=10$ & 0.1069 & 0.2471 & 0.6794 \\
& $\alpha=100$ & 0.1090 & 0.2468 & 0.6869 \\

\bottomrule
\end{tabular}
\end{table}

\begin{table}[ht!]  
\caption{ Downstream task performance comparison under different $\alpha$ values. (a) Forecasting error
metrics (lower is better). (b) Classification performance metrics (higher is better).}
\label{tab:downtrema_performance}
\begin{subtable}{0.48\linewidth}
    \centering
    \caption{}  \label{tab:downstream_alpha_forecasting}
    \begin{tabular}{l S[table-format=1.4] S[table-format=1.4]}
    \toprule
    \textbf{Example} & {\textbf{MSE↓}} & {\textbf{MAE↓}} \\
    \midrule
    $\alpha=1$   & 1.7030 & 0.9112 \\
    $\alpha=10$  & 1.7230 & 0.9140 \\
    $\alpha=100$ & 1.7410 & 0.9177 \\
    \bottomrule
    \end{tabular}
\end{subtable}
\hfill 
\begin{subtable}{0.48\linewidth}
    \centering
    \caption{Sensitivity to $\alpha$}
    \label{tab:downstream_alpha_classification}
    \begin{tabular}{l  S[table-format=2.2] S[table-format=2.2] S[table-format=2.2] S[table-format=2.2] S[table-format=2.2]}
    \toprule
    \textbf{Example} & {\textbf{Acc.↑}} & {\textbf{Prec.↑}} & {\textbf{Rec.↑}} & {\textbf{F1↑}} & {\textbf{Avg↑}}\\  
    \midrule
    $\alpha = 1$ & 92.38 & 94.67 & 84.04 & 87.82 & 89.73 \\
    $\alpha = 10$ & 93.09 & 94.56 & 83.44 & 87.65 & 89.69 \\
    $\alpha = 100$ & 93.28 & 94.16 & 84.30 & 88.15 & 89.97 \\

    \bottomrule
    \end{tabular}
\end{subtable}

\end{table}


\subsection{Hyperparameters}
\label{app:hyperparameters}
Preprocessing. We apply z-score normalization using a StandardScaler fit only on the training split, and reuse its statistics to transform validation and test splits (preventing leakage). For multivariate inputs, we normalize per channel. For UEA datasets we follow the provided Normalizer. All normalization is applied before computing reconstruction losses.

The following tables summarize the key hyperparameters used for our pre-training and fine-tuning experiments.

\subsubsection{Forecasting Hyperparameters}

\begin{table}[H]
\centering
\begin{minipage}{0.48\linewidth}
    \caption{Pre-training}
    \label{tab:hyper_pre_forecast}
    \centering
    \begin{tabular}{ll}
    \toprule
    \textbf{Parameter} & \textbf{Value} \\
    \midrule
    Sequence Length & 96 \\
    Learning rate   & \num{1e-4} \\
    Batch size      & 16 \\
    Epochs          & 50 \\
    Alpha (for SDSC) & 10 \\
    Optimizer       & Adam \\
    \bottomrule
    \end{tabular}
\end{minipage}
\hfill 
\begin{minipage}{0.48\linewidth}
    \caption{Fine-tuning}
    \label{tab:hyper_fine_forecast}
    \centering
    \begin{tabular}{ll}
    \toprule
    \textbf{Parameter} & \textbf{Value} \\
    \midrule
    Sequence Length & 96 \\
    Learning rate   & \num{1e-4} \\
    Batch size      & 16 \\
    Epochs          & 10 \\
    Alpha (for SDSC) & 10 \\
    Optimizer       & Adam \\
    \bottomrule
    \end{tabular}
\end{minipage}
\end{table}

\subsubsection{Classification Hyperparameters}

\begin{table}[H]
\centering
\begin{minipage}{0.48\linewidth}
    \caption{Pre-training}
    \label{tab:hyper_pre_class}
    \centering
    \begin{tabular}{ll}
    \toprule
    \textbf{Parameter} & \textbf{Value} \\
    \midrule
    Learning rate   & \num{1e-4} \\
    Batch size      & 32 \\
    Epochs          & 20 \\
    Alpha (for SDSC) & 10 \\
    Optimizer       & Adam \\
    \bottomrule
    \end{tabular}
\end{minipage}
\hfill
\begin{minipage}{0.48\linewidth}
    \caption{Fine-tuning}
    \label{tab:hyper_fine_class}
    \centering
    \begin{tabular}{ll}
    \toprule
    \textbf{Parameter} & \textbf{Value} \\
    \midrule
    Learning rate   & \num{1e-4}/\num{3e-4} \\
    Batch size      & 32 \\
    Epochs          & 100/300 \\
    Alpha (for SDSC) & 10 \\
    Optimizer       & Adam \\
    \bottomrule
    \end{tabular}
\end{minipage}
\end{table}

\subsection{Full Pre-training Forecasting Results}
\label{app:full_pre_forecasting_results}

\begin{table}[h]
\caption{Full pre-training reconstruction performance for forecasting datasets.}
\label{tab:pre_forecasting_full}
\centering
\small 
\begin{tabular}{ll S[table-format=3.4] S[table-format=1.4] S[table-format=1.4]}
\toprule
\textbf{Dataset} & \textbf{Loss} & {MSE↓} & {MAE↓}  & {SDSC↑} \\
\midrule
\multirow{5}{*}{ETTh1} 
 & MSE    & 1.1980 & 0.6863  & 0.5573 \\
 & SoftDTW & 1.9270 & 0.9424  & 0.4199\\
 & PCC    & 2.4280 & 1.0950  & 0.4051 \\
 & SI-SNR & 221.9000 & 11.6300  & 0.0713 \\
\cmidrule(lr){2-5}
 & \textbf{SDSC}   & 1.5430 & 0.7653  & 0.5725 \\
 & \textbf{Hybrid} & 1.2070 & 0.6738  & 0.5774 \\
\midrule
\multirow{5}{*}{ETTh2}
 & MSE    & 0.6229 & 0.4595 & 0.7263 \\
 & SoftDTW & 0.9799& 0.6386 & 0.6229 \\
 & PCC    & 0.9711 & 0.6350 & 0.6190 \\
 & SI-SNR & 0.9586 & 0.6263 & 0.6242\\
\cmidrule(lr){2-5}
 & \textbf{SDSC}   & 0.6553 & 0.4643 & 0.7206 \\
 & \textbf{Hybrid} & 0.6210 & 0.4487 & 0.7320 \\
\midrule
 \multirow{5}{*}{ETTm1}
 & MSE    & 1.1740 & 0.6532 & 0.5759 \\
 & SoftDTW & 1.8140 & 0.8955  & 0.4334 \\
 & PCC    & 2.6570 & 1.1030 & 0.4149 \\
 & SI-SNR & 1.7990 & 0.9406 & 0.3999 \\
\cmidrule(lr){2-5}
 & \textbf{SDSC}   & 1.7880 & 0.7975 & 0.5838 \\
 & \textbf{Hybrid} & 1.1780 & 0.6420 & 0.5886 \\
\midrule
 \multirow{5}{*}{ETTm2}
 & MSE    & 0.0269 & 0.1121 & 0.9355  \\
 & SoftDTW & 0.7497 & 0.5150  & 0.7078 \\
 & PCC    & 0.7235 & 0.5016 & 0.7105\\
 & SI-SNR & 0.7413 & 0.4978 & 0.7155 \\
\cmidrule(lr){2-5}
 & \textbf{SDSC}   & 0.0280 & 0.1138 & 0.9345 \\
 & \textbf{Hybrid} & 0.0242 & 0.1052 & 0.9395 \\
\midrule
 \multirow{5}{*}{Weather}
 & MSE    & 0.0863 & 0.1126 & 0.9022 \\
 & SoftDTW & 0.7056 & 0.4354  & 0.5886 \\
 & PCC    & 0.6908 & 0.4337 & 0.5970 \\
 & SI-SNR & 0.6848 & 0.4149 & 0.6189 \\
\cmidrule(lr){2-5}
 & \textbf{SDSC}   & 0.1024 & 0.1045 & 0.9097 \\
 & \textbf{Hybrid} & 0.0806 & 0.1007 & 0.9132 \\
\midrule
 \multirow{5}{*}{Electricity}
 & MSE    & 0.0970 & 0.2136 & 0.8544 \\
 & SoftDTW & 1.1060 & 0.7871  & 0.4252 \\
 & PCC    & 0.7070 & 0.6366 & 0.5180 \\
 & SI-SNR & 1.0260 & 0.6743 & 0.5398 \\
\cmidrule(lr){2-5}
 & \textbf{SDSC}   & 0.1039 & 0.2215 & 0.8524 \\
 & \textbf{Hybrid} & 0.0842 & 0.1960 & 0.8679 \\
\midrule
 \multirow{5}{*}{Traffic}
 & MSE    & 0.1914 & 0.2304 & 0.8382  \\
 & SoftDTW & 2.0090 & 0.9881  & 0.2949 \\
 & PCC    & 1.1250 & 0.7226 & 0.4270 \\
 & SI-SNR & 17.2500 & 3.0020  & 0.1964 \\
\cmidrule(lr){2-5}
 & \textbf{SDSC}   & 0.2229 & 0.2418 & 0.8327  \\
 & \textbf{Hybrid} & 0.1529 & 0.1914 & 0.8700 \\
\midrule
\midrule
\multirow{5}{*}{\textbf{Avg (Forecasting)}} 
 & MSE    & 0.4852 & 0.3525 & 0.7670\\
 & SoftDTW & 1.3273 & 0.7432  & 0.4990 \\
 & PCC    & 1.3289 & 0.6705 & 0.5274 \\
 & SI-SNR & 34.9085& 2.5408 & 0.4523 \\
\cmidrule(lr){2-5}
 & \textbf{SDSC}   & 0.6348 & 0.3870  & 0.7723\\
 & \textbf{Hybrid} & \textbf{0.4783} & \textbf{0.3368} & \textbf{0.7841} \\
\bottomrule
\end{tabular}
\end{table}
Note: SI-SNR values are reported for completeness, but they are on a different numerical scale than MSE/MAE/SDSC. 
In some datasets (e.g., ETTh1), training with SI-SNR fails to converge, leading to unstable or very large values. 
These results should therefore be interpreted with caution (see main text for discussion).

\clearpage
\subsection{Full \texorpdfstring{$\lambda$}{lambda}=0.5 Pre-training Forecasting Results}
\label{app:full_pre_0.5_forecasting_results}

\begin{table}[h]
\caption{Full pre-training reconstruction performance for forecasting datasets with fixed $\lambda = 0.5$}
\label{tab:pre_0.5_forecasting_full}
\centering
\small 
\begin{tabular}{ll  S[table-format=1.4] S[table-format=1.4]}
\toprule
\textbf{Dataset} &  {MSE↓} & {MAE↓}  & {SDSC↑} \\
\midrule
ETTh1 & 1.7480 & 0.9185 & 0.3997\\
\midrule
ETTh2 & 0.9359 & 0.6218 & 0.6262\\
\midrule
ETTm1 & 1.8250 & 0.9354 & 0.4026\\
\midrule
ETTm2 & 0.7236 & 0.5068 & 0.7060\\
\midrule
Weather & 0.6919 & 0.4449 & 0.5721\\
\midrule
Electricity & 0.7146 & 0.6547 & 0.4703\\
\midrule
Traffic & 0.9691 & 0.6406 & 0.4929\\
\midrule
\textbf{Avg (Forecasting)} & 1.0869 & 0.6747 & 0.5243\\
\bottomrule
\end{tabular}
\end{table}

\subsection{Full Pre-training Classification Results}
\label{app:full_pre_classification_results} 
\begin{table}[ht!]
\caption{Full pre-training reconstruction performance for classification datasets.}
\label{tab:pre_classification_full}
\centering
\small 
\begin{tabular}{ll S[table-format=3.4] S[table-format=1.4] S[table-format=1.4]}
\toprule
\textbf{Dataset} & \textbf{Loss} & {MSE↓} & {MAE↓} & {SDSC↑} \\
\midrule
\multirow{5}{*}{Epilepsy} 
 & MSE    & 0.0807 & 0.2456  & 0.6244  \\
 & Soft-DTW & \textbf{0.0719} & \textbf{0.2384} & 0.6407  \\
 & PCC    & 0.3128 & 0.4543  & 0.2468 \\
 & SI-SNR & 0.3235 & 0.4612  & 0.2435 \\
\cmidrule(lr){2-5}
 & \textbf{SDSC(Ours)}   & 0.1300 & 0.2740  & 0.6736 \\
 & \textbf{Hybrid(Ours)} & 0.1099 & 0.2484  & \textbf{0.6856} \\
\midrule
\multirow{5}{*}{SleepEEG}
 & MSE    & 100.5599 & 6.8082 & 0.5966 \\
 & Soft-DTW & \textbf{98.1958} & \textbf{6.7118} & 0.6012\\
 & PCC    & 239.7082 & 8.5638 & 0.0775 \\
 & SI-SNR & 236.8986 & 8.5080 & 0.0950 \\
\cmidrule(lr){2-5}
 & \textbf{SDSC(Ours)}   & 147.9205 & 7.4512 & \textbf{0.6483}  \\
 & \textbf{Hybrid(Ours)} & 100.5852 & 6.8087 & 0.6105 \\
\midrule
\midrule
\multirow{5}{*}{\textbf{Avg (Classification)}} 
 & MSE    & 50.3203 & 3.5269 & 0.6105 \\
 & Soft-DTW & \textbf{49.1339} & \textbf{3.4751} & 0.6210  \\
 & PCC    & 120.0105 & 4.5091 & 0.1622 \\
 & SI-SNR & 118.6110 & 4.4846 & 0.1693 \\
\cmidrule(lr){2-5}
 & \textbf{SDSC(Ours)}   & 74.0253 & 3.8626 & \textbf{0.6610}\\
 & \textbf{Hybrid(Ours)} & 50.3471 & 3.5286 & 0.6481 \\
\bottomrule
\end{tabular}
\end{table}

\clearpage
\subsection{Full \texorpdfstring{$\lambda$}{lambda}=0.5 Pre-training Classification Results}
\label{app:full_pre_0.5_classification_results}

\begin{table}[h]
\caption{Full pre-training reconstruction performance for classification datasets with fixed $\lambda = 0.5$}
\label{tab:pre_0.5_classification_full}
\centering
\small 
\begin{tabular}{ll S[table-format=1.4] S[table-format=1.4] S[table-format=1.4]}
\toprule
\textbf{Dataset} &   {MSE↓} & {MAE↓}  & {SDSC↑} \\
\midrule
Epilepsy & 0.1096 & 0.2470 & 0.6881 \\
\midrule
SleepEEG & 110.3683 & 7.1989 & 0.5623 \\
\midrule
\textbf{Avg (Classification)} & 55.2390 & 3.7230 & 0.6252 \\
\bottomrule
\end{tabular}
\end{table}

\clearpage
\subsection{Full Forecasting Result}
\label{app:full_forecasting_results}

\begin{table}[ht!]
\caption{Forecasting fine-tuning performance with all baselines.}
\centering
\small 

\begin{tabular}{ll S[table-format=1.3] S[table-format=1.3]}
\toprule
\textbf{Dataset} & \textbf{Pre-training Loss} & {\textbf{MSE↓}} & {\textbf{MAE↓}} \\
\midrule
\multirow{5}{*}{ETTh1} 
 & MSE    & 0.380 & 0.408  \\
 & Soft-DTW & 0.384 & 0.410 \\
 & PCC    & 0.382 & 0.408 \\ 
 & SI-SNR & 0.381 & \textbf{0.399} \\ 
\cmidrule(lr){2-4}
 & \textbf{SDSC(Ours)}   & \textbf{0.379} & 0.406  \\
 & \textbf{Hybrid(Ours)} & 0.382 & 0.406\\
\midrule
\multirow{5}{*}{ETTh2}
 & MSE    & 0.304 & 0.350  \\
 & Soft-DTW & 0.306 & 0.351 \\
 & PCC    & 0.304 & 0.350 \\
 & SI-SNR & \textbf{0.303} & \textbf{0.349} \\ 
\cmidrule(lr){2-4}
 & \textbf{SDSC(Ours)}   & 0.306 & 0.352 \\
 & \textbf{Hybrid(Ours)} & 0.304 & 0.350 \\
\midrule
\multirow{5}{*}{ETTm1}
 & MSE    & 0.327 & \textbf{0.363}  \\
 & Soft-DTW & 0.327 & 0.365 \\
 & PCC    & 0.321 & 0.364 \\
 & SI-SNR & 0.345 & 0.373 \\
\cmidrule(lr){2-4}
 & \textbf{SDSC(Ours)}    &\textbf{0.324} & 0.364 \\
 & \textbf{Hybrid(Ours)} & 0.325 & 0.364 \\
\midrule
\multirow{5}{*}{ETTm2}
 & MSE    & \textbf{0.185} & \textbf{0.275}\\
 & Soft-DTW & 0.190 & 0.278\\
 & PCC    & 0.194 & 0.279 \\
 & SI-SNR & 0.189 & 0.279 \\
\cmidrule(lr){2-4}
 & \textbf{SDSC(Ours)}   & 0.191 & 0.278\\
 & \textbf{Hybrid(Ours)} & 0.188  & 0.276\\
 \midrule
\multirow{5}{*}{Weather}
 & MSE     & 0.169 & \textbf{0.213}  \\
 & Soft-DTW & 0.176 & 0.220\\
 & PCC    & \textbf{0.168} & 0.215\\
 & SI-SNR & 0.196 & 0.239 \\
\cmidrule(lr){2-4}
 & \textbf{SDSC(Ours)}   & \textbf{0.168} & \textbf{0.213} \\
 & \textbf{Hybrid(Ours)} & 0.169 & 0.215 \\
\midrule
\multirow{5}{*}{Electricity}
 & MSE    & 0.200 & 0.291 \\
 & Soft-DTW & 0.200 & 0.290 \\
 & PCC    & 0.199 & \textbf{0.288} \\
 & SI-SNR & 0.203 & 0.292 \\
\cmidrule(lr){2-4}
 & \textbf{SDSC(Ours)}   & 0.200 & 0.293 \\
 & \textbf{Hybrid(Ours)} & \textbf{0.198} & \textbf{0.288} \\
\midrule
\multirow{5}{*}{Traffic}
 & MSE    & 0.497 & 0.312  \\
 & Soft-DTW & 0.535 & 0.337 \\
 & PCC    & 0.505 & 0.319 \\
 & SI-SNR & 0.550 & 0.345 \\
\cmidrule(lr){2-4}
 & \textbf{SDSC(Ours)}   & \textbf{0.492} & \textbf{0.309} \\
 & \textbf{Hybrid(Ours)} & 0.494  & 0.315\\
\midrule
\midrule
\multirow{5}{*}{Avg}
 & MSE    & 0.295 & \textbf{0.316} \\
 & Soft-DTW & 0.303 & 0.322 \\
 & PCC    & 0.296 & 0.318 \\
 & SI-SNR & 0.310 & 0.325 \\
\cmidrule(lr){2-4}
 & \textbf{SDSC(Ours)}   & \textbf{0.294} & \textbf{0.316} \\
 & \textbf{Hybrid(Ours)} & \textbf{0.294} & \textbf{0.316} \\
\bottomrule
\end{tabular}
\end{table}

\clearpage
\subsection{Full \texorpdfstring{$\lambda$}{lambda}=0.5 Forecasting Results}
\label{app:full_0.5_forecasting_results}

\begin{table}[h]
\caption{ Forecasting fine-tuning performance with all baselines with fixed $\lambda = 0.5$}
\label{tab:0.5_forecasting_full}
\centering
\small 
\begin{tabular}{ll  S[table-format=1.4] S[table-format=1.4]}
\toprule
\textbf{Dataset} &  {MSE↓} & {MAE↓} \\
\midrule
ETTh1 & 0.380 & 0.405 \\
\midrule
ETTh2 & 0.307 & 0.352 \\    
\midrule
ETTm1 & 0.319 & 0.362 \\
\midrule
ETTm2 & 0.193 & 0.279 \\
\midrule
Weather & 0.168 & 0.215 \\
\midrule
Electricity & 0.200 & 0.289 \\
\midrule
Traffic & 0.506 & 0.318 \\
\midrule
\textbf{Avg (Forecasting)} & 0.2961 & 0.3171\\
\bottomrule
\end{tabular}
\end{table}

\clearpage
\subsection{Full Freeze Classification Result}
\label{app:full_freeze_classification_results}
\begin{table}[H]
\caption{Full freeze classification performance across all scenarios. Full results for all datasets are presented to ensure reproducibility.}\centering
\small
\sisetup{detect-weight=true, detect-family=true}

\begin{tabular}{c c l S[table-format=2.2] S[table-format=2.2] S[table-format=2.2] S[table-format=2.2] S[table-format=2.2]}
\toprule
\multicolumn{2}{c}{\textbf{Scenario}} & \textbf{Loss} & {\textbf{Acc.↑}} & {\textbf{Prec.↑}} & {\textbf{Rec.↑}} & {\textbf{F1↑}} & {\textbf{Avg↑}} \\
\midrule
\multirow{10}{*}{\textbf{In Domain}} & \multirow{5}{*}{\makecell{Epilepsy \\ $\downarrow$ \\ Epilepsy}}
 & MSE      & 90.69 & 93.23 & 77.23 & 82.26 & 85.85 \\
 & & Soft-DTW & 80.21 & 40.11 & 50.00 & 44.51 & 53.71 \\
 & & PCC    & 80.21 & 40.10 & 50.00 & 44.51 & 53.71 \\
 & & SI-SNR & 80.21 & 40.10 & 50.00 & 44.51 & 53.71  \\
\cmidrule(lr){3-8}
 & & \textbf{SDSC(Ours)}   & 91.86 & 93.82 & 80.27 & 84.98 & 87.73 \\
 & & \textbf{Hybrid(Ours)} & \textbf{92.44} & \textbf{93.58} & \textbf{82.13} & \textbf{86.38} & \textbf{88.63} \\
\cmidrule(lr){2-8}
 & \multirow{5}{*}{\makecell{SleepEEG \\ $\downarrow$ \\ SleepEEG}}
 & MSE    & 60.20 & 52.90 & 49.75 & 46.92 & 52.44 \\
 & & Soft-DTW & 57.31 & 41.67 & 44.72 & 42.52 & 46.56 \\
 & & PCC    & \textbf{62.59} & 54.65 & 53.12 & 48.88 & 54.81 \\
 & & SI-SNR & 61.95 & \textbf{54.69} & \textbf{53.62} & \textbf{49.30} & \textbf{54.89} \\
 \cmidrule(lr){3-8}
 & & \textbf{SDSC(Ours)}   & 60.89 & 54.55 & 49.59 & 46.72 & 52.94 \\
 & & \textbf{Hybrid(Ours)} & 60.01 & 51.86 & 49.09 & 46.55 & 51.88 \\
\midrule
\multirow{20}{*}{\textbf{Cross Domain}} & \multirow{5}{*}{\makecell{SleepEEG \\ $\downarrow$ \\ Epilepsy}}
 & MSE    & 80.21 & 40.11 & 50.00 & 44.51 & 53.71 \\
 & & Soft-DTW & 80.21 & 40.11 & 50.00 & 44.51 & 53.71 \\
 & & PCC    & 80.21 & 40.11 & 50.00 & 44.51 & 53.71 \\
 & & SI-SNR  & 80.21 & 40.11 & 50.00 & 44.51 & 53.71  \\
 \cmidrule(lr){3-8}
 & & \textbf{SDSC(Ours)}   & 80.21 & 40.11 & 50.00 & 44.51 & 53.71 \\
 & & \textbf{Hybrid(Ours)} & 80.21 & 40.11 & 50.00 & 44.51 & 53.71 \\

 \cmidrule(lr){2-8}
 & \multirow{5}{*}{\makecell{SleepEEG \\ $\downarrow$ \\ FD-B}}
 & MSE      & \textbf{52.19} & 35.80 & 38.21 & 33.87 & 40.02 \\
 & & Soft-DTW & 52.30 & 37.99 & \textbf{38.49} & \textbf{36.06} & \textbf{41.21} \\
 & & PCC    & 52.00 & 37.64 & 38.06 & 32.29 & 39.99 \\
 & & SI-SNR & 49.79 & \textbf{39.44} & 36.45 & 28.84 & 38.63 \\
\cmidrule(lr){3-8}
 & & \textbf{SDSC(Ours)} &  51.71 & 37.11 & 37.86 & 31.84 & 39.63\\
 & & \textbf{Hybrid(Ours)} & 51.43 & 34.34 & 37.65 & 34.01 &  39.96 \\

 \cmidrule(lr){2-8}
 & \multirow{5}{*}{\makecell{SleepEEG \\ $\downarrow$ \\ Gesture}}
 & MSE      & \textbf{70.00} & 64.80 & \textbf{70.00} & \textbf{66.17} & 67.74 \\
 & &Soft-DTW & 69.17 & \textbf{72.31} & 69.16 & 65.22 & \textbf{68.97} \\
 & & PCC    & 63.33 & 63.68 & 63.33 & 58.84 & 62.30 \\
 & & SI-SNR & 64.16 & 59.18 & 64.16 & 58.92 & 61.61 \\
\cmidrule(lr){3-8}
 & & \textbf{SDSC(Ours)}      & 68.33 & \textbf{66.34} & 68.33 & 63.89 & 66.72\\
 & & \textbf{Hybrid(Ours)}  & 70.00 & 66.09 & \textbf{70.00} & \textbf{66.17} & 68.07 \\

 \cmidrule(lr){2-8}
 & \multirow{5}{*}{\makecell{SleepEEG \\ $\downarrow$ \\ EMG}}
 & MSE    & 46.34 & 15.45 & 33.33 & 21.11 & 29.06 \\
 & &Soft-DTW & 46.34 & 15.45 & 33.33 & 21.11 & 29.00 \\
 & & PCC    & 46.34 & 15.45 & 33.33 & 21.11 & 29.06 \\
 & & SI-SNR  & 46.34 & 15.45 & 33.33 & 21.11 & 29.06 \\
  \cmidrule(lr){3-8}
 & & \textbf{SDSC(Ours)}     & 46.34 & 15.45 & 33.33 & 21.11 & 29.06 \\
 & & \textbf{Hybrid(Ours)}  & 46.34 & 15.45 & 33.33 & 21.11 & 29.06 \\
\bottomrule
\end{tabular}
\end{table}

In some cross-domain cases (e.g., SleepEEG→Epilepsy, SleepEEG→EMG), the reported metrics are identical across losses. This is not due to code reuse but rather because the small dataset failed to converge under all objectives, leading to degenerate identical predictions. We verified by running independent trials.

\clearpage
\subsection{Full Classification Result}
\label{app:full_classification_results}
\begin{table}[H]
\caption{Full fine-tuning classification performance across all scenarios.}
\label{tab:app_finetuning_classification_full}
\centering
\small
\sisetup{detect-weight=true, detect-family=true}

\begin{tabular}{c c l S[table-format=2.2] S[table-format=2.2] S[table-format=2.2] S[table-format=2.2] S[table-format=2.2]}
\toprule
\multicolumn{2}{c}{\textbf{Scenario}} & \textbf{Loss} & {\textbf{Acc.↑}} & {\textbf{Prec.↑}} & {\textbf{Rec.↑}} & {\textbf{F1↑}} & {\textbf{Avg↑}} \\
\midrule
\multirow{10}{*}{\textbf{In Domain}} & \multirow{5}{*}{\makecell{Epilepsy \\ $\downarrow$ \\ Epilepsy}}
 & MSE      &  \textbf{94.20} & \textbf{94.64} & \textbf{86.77} & \textbf{90.02} & \textbf{91.40} \\
 & & Soft-DTW & 93.08 & 94.16 & 83.69 & 87.71 & 89.66 \\
 & & PCC    & 94.07 & \textbf{94.64} & 86.34 & 89.74 & 91.20 \\
 & & SI-SNR & 93.15 & 94.18 & 83.88 & 87.86 & 89.78 \\
 \cmidrule(lr){3-8}
 & & \textbf{SDSC(Ours)}    &  93.97 & 94.46 & 86.16 & 89.56 & 91.04 \\
 & & \textbf{Hybrid(Ours)}  & 93.91 & 94.59 & 85.89 & 89.42 & 90.95 \\
\cmidrule(lr){2-8}
 & \multirow{5}{*}{\makecell{SleepEEG \\ $\downarrow$ \\ SleepEEG}}
 & MSE      & 65.11 & 57.10 & 55.70 & 52.16 & 57.52 \\
 & & Soft-DTW & 65.05 & 57.19 & 56.32 & 51.15 & 57.43\\
 & & PCC    & 65.44 & \textbf{57.69} & \textbf{56.34} & 52.68 & 58.04 \\
 & & SI-SNR & \textbf{65.46} & 57.51 & 56.66 & \textbf{52.75} & \textbf{58.09} \\
\cmidrule(lr){3-8}
 & & \textbf{SDSC(Ours)}   & 65.22 & 57.55 & 54.92 & 51.81 & 57.38 \\
 & & \textbf{Hybrid(Ours)} & 65.13 & 56.62 & 56.30 & 51.00 & 57.26 \\
\midrule
\multirow{20}{*}{\textbf{Cross Domain}}& \multirow{5}{*}{\makecell[c]{SleepEEG\\$\downarrow$\\Epilepsy}}
  & MSE      & 95.19 & 94.30 & 90.20 & 92.07 & 92.94 \\
  & & Soft-DTW & \textbf{95.40} & 94.85 & \textbf{90.35} & \textbf{92.39} & \textbf{93.25} \\
  & & PCC    & 91.35 & 94.25 & 78.57 & 83.65 & 86.96 \\
  & & SI-SNR & 94.16 & \textbf{95.24} & 86.16 & 89.82 & 91.35 \\
  \cmidrule(lr){3-8}
  & & \textbf{SDSC(Ours)}       & 92.03 & 94.33 & 80.48 & 85.27 & 88.03\\
  & & \textbf{Hybrid(Ours)}  & \textbf{95.19} & 94.30 & \textbf{90.20} & \textbf{92.07} & \textbf{92.94} \\

\cmidrule(lr){2-8}
& \multirow{5}{*}{\makecell[c]{SleepEEG\\$\downarrow$\\FD-B}}
  & MSE      &  63.88 & 69.36& 72.98&  69.98& 69.05\\
  & & Soft-DTW & 60.68 & 65.76 & 70.64 & 67.68 & 66.19 \\
  & & PCC    & \textbf{66.24} & \textbf{73.78} & 72.88 & 71.85 & \textbf{71.24} \\
  & & SI-SNR & 65.35 & 65.10 & 72.30 & 66.48 & 67.31 \\
  \cmidrule(lr){3-8}
  & & \textbf{SDSC(Ours)}  &  65.94 & 72.33 & 73.45 & \textbf{72.29}  & 71.00 \\
  & & \textbf{Hybrid(Ours)}  & 64.26 & 68.02 & \textbf{73.50} & 70.06 & 68.96 \\

\cmidrule(lr){2-8}
& \multirow{5}{*}{\makecell[c]{SleepEEG\\$\downarrow$\\Gesture}}
  & MSE      & 78.33 & 79.85 & 78.33 & 77.13 & 78.41 \\
 & & Soft-DTW & 79.17 & 78.65 & 79.17 & 77.35 & 78.58 \\
  & & PCC    & \textbf{80.00} & 80.77 & \textbf{80.00} & 78.47 & 79.81 \\
  & & SI-SNR & \textbf{80.00} & 80.33 & \textbf{80.00} & 78.31 & 79.66 \\
  \cmidrule(lr){3-8}
  & & \textbf{SDSC(Ours)}   & \textbf{80.00} & 80.21 & \textbf{80.00} & \textbf{79.32}  & \textbf{79.88}\\
  & & \textbf{Hybrid(Ours)} & 78.33 & \textbf{81.20} & 78.33 & 76.15 & 78.50 \\
 
\cmidrule(lr){2-8}
& \multirow{5}{*}{\makecell[c]{SleepEEG\\$\downarrow$\\EMG}}
  & MSE     & \textbf{97.56} & \textbf{98.33} & \textbf{98.04} & \textbf{98.14} & \textbf{98.18} \\
 & & Soft-DTW & \textbf{97.56} & \textbf{98.33} & \textbf{98.04} & \textbf{98.14} & \textbf{98.18} \\
  & & PCC    & \textbf{97.56} & \textbf{98.33} & \textbf{98.04} & \textbf{98.14} & \textbf{98.18} \\
  & & SI-SNR & \textbf{97.56} & \textbf{98.33} & \textbf{98.04} & \textbf{98.14} & \textbf{98.18} \\
  \cmidrule(lr){3-8}
  & & \textbf{SDSC(Ours)}      & 95.12 & 96.83 &91.37 & 93.62 & 94.24\\
  & & \textbf{Hybrid(Ours)}  & 95.12 & 96.83 & 91.37 & 93.62 & 94.24 \\
  
\bottomrule
\end{tabular}
\end{table}

\clearpage
\subsection{Full \texorpdfstring{$\lambda$}{lambda}=0.5 Pre-training Classification Results}
\label{app:full_0.5_classification_results}
\begin{table}[H]
\caption{ Full freeze classification performance across all scenarios with fixed $\lambda = 0.5$. Full results
for all datasets are presented to ensure reproducibility.}
\label{tab:0.5_classification_full}
\centering
\small
\sisetup{detect-weight=true, detect-family=true}

\begin{tabular}{c c  S[table-format=2.2] S[table-format=2.2] S[table-format=2.2] S[table-format=2.2] S[table-format=2.2]}
\toprule
\multicolumn{2}{c}{\textbf{Scenario}} & {\textbf{Acc.↑}} & {\textbf{Prec.↑}} & {\textbf{Rec.↑}} & {\textbf{F1↑}} & {\textbf{Avg↑}} \\
\midrule
\multirow{5}{*}{\textbf{In Domain}} 
& \makecell{Epilepsy \\ $\downarrow$ \\ Epilepsy} &80.21 & 40.11 & 50.00 & 44.51 & 46.21\\
\cmidrule(lr){2-7}
& \makecell{SleepEEG \\ $\downarrow$ \\ SleepEEG} &58.78 & 50.94 & 46.69 & 44.31 & 50.18\\
\midrule
\multirow{11}{*}{\textbf{Cross Domain}}& 
\makecell[c]{SleepEEG\\$\downarrow$\\Epilepsy} & 80.21 & 40.11 & 50.00 & 44.51 & 46.21\\
\cmidrule(lr){2-7}
& \makecell[c]{SleepEEG\\$\downarrow$\\FD-B} &   50.02 & 36.61 & 38.09 & 34.51 & 39.81\\
\cmidrule(lr){2-7}
& \makecell[c]{SleepEEG\\$\downarrow$\\Gesture} & 68.33 & 68.03 & 68.33 & 64.60 & 67.32\\
\cmidrule(lr){2-7}
& \makecell[c]{SleepEEG\\$\downarrow$\\EMG} & 46.34 & 15.45 & 33.33 & 21.11 & 29.06\\
\bottomrule
\end{tabular}
\end{table}

\clearpage

\subsection{Practical Guideline for Loss Selection}
\label{app:pratical_guidline}
\begin{table}[h]
\centering
\begin{tabular}{p{0.22\linewidth}p{0.38\linewidth}p{0.30\linewidth}}
\toprule
\textbf{Loss Type} & \textbf{Recommended Usage Scenario} & \textbf{Example Dataset / Task} \\ 
\midrule
$\mathcal{L}_{\text{MSE}}$ & Amplitude-critical tasks requiring precise numeric matching; performs well when signal magnitude is stable across domains. & FD-B (Cross-domain forecasting) \\
$\mathcal{L}_{\text{SDSC}}$ & Structure-critical tasks emphasizing polarity and waveform shape consistency; suitable for oscillatory or physiological signals. & SleepEEG (In-domain SSL pretraining) \\
Hybrid ($\lambda_{sdsc}$,$\lambda_{mse}$) & Balanced reconstruction for tasks with mixed structural and amplitude sensitivity; provides stable representation across varied regimes. & Epilepsy EEG (Hybrid reconstruction and fine-tuning) \\
\bottomrule
\end{tabular}
\caption{Guidelines for selecting between MSE, SDSC, and the hybrid loss across different scenarios and datasets.}
\end{table}

\subsection{The Use of Large Language Models}
We utilized a large language model (e.g., Google's Gemini, OpenAI's GPT-4) to assist in polishing the writing of the manuscript. Its role was limited to improving clarity, refining grammar, and ensuring consistent terminology. The core ideas, experimental design, and all results and analyses presented in this paper are entirely our own.

\clearpage
\section{Lemma}
\label{app:proof_of_lemma}

\begin{lemma}[Boundedness of SDSC]
\label{lem:bounded_sdsc}
For any two discrete signals $E = \{E(s)\}_{s \in S}$ and $R = \{R(s)\}_{s \in S}$, the Signal Dice Similarity Coefficient, $\text{SDSC}(E, R)$, is bounded such that $0 \le \text{SDSC} \le 1$.
\end{lemma}

\begin{proof}
We begin with the definition of the discrete SDSC:
\[
\mathrm{SDSC}(E(t),R(t)) \approx 
\frac{2 \cdot \sum_{s \in S} \bigl(H(E(s)R(s)) \cdot M(s)\bigr)}
{\sum_{s \in S}\bigl(|E(s)|+|R(s)|\bigr)+\epsilon}.
\]
Here, $H(\cdot)$ is the Heaviside step function with the convention $H(0)=0$, $S$ is the set of discrete sampling points with $S \subset T$, and $\epsilon>0$ is a small constant to prevent division by zero.

\paragraph{Lower Bound ($\mathrm{SDSC}\ge 0$).}
For each sampling point $s \in S$:
\begin{itemize}
    \item $H(E(s)R(s)) \in \{0,1\}$, hence it is non-negative.
    \item $M(s)=\min\{|E(s)|,|R(s)|\}$ is non-negative.
    \item $|E(s)|$ and $|R(s)|$ are non-negative.
\end{itemize}
Therefore, both the numerator and the denominator are non-negative, implying $\mathrm{SDSC}\ge 0$.

\paragraph{Upper Bound ($\mathrm{SDSC}\le 1$).}
Let
\[
N_s = 2 \cdot H(E(s)R(s)) \cdot M(s),
\qquad
D_s = |E(s)| + |R(s)|.
\]
We show $N_s \le D_s$ for every $s \in S$.

\begin{itemize}
    \item \textbf{Case 1: Strictly same sign ($E(s)R(s) > 0$).}
    In this case, $H(E(s)R(s))=1$, hence $N_s = 2M(s)$. For any $a,b\ge 0$,
    $2\min(a,b) \le a+b$. Applying this with $a=|E(s)|$ and $b=|R(s)|$ yields
    \[
    N_s = 2\min\{|E(s)|,|R(s)|\} \le |E(s)|+|R(s)| = D_s.
    \]

    \item \textbf{Case 2: Opposite signs or one is zero ($E(s)R(s)\le 0$).}
    By the convention $H(0)=0$, we have $H(E(s)R(s))=0$, so $N_s=0$.
    Since $D_s=|E(s)|+|R(s)|\ge 0$, it follows that $N_s \le D_s$.
\end{itemize}

Summing $N_s \le D_s$ over $s\in S$ gives
\[
\sum_{s\in S} 2\,H(E(s)R(s))\,M(s)
\le \sum_{s\in S}\bigl(|E(s)|+|R(s)|\bigr)
\le \sum_{s\in S}\bigl(|E(s)|+|R(s)|\bigr) + \epsilon.
\]
Dividing by the positive denominator $\sum_{s\in S}(|E(s)|+|R(s)|)+\epsilon$ yields
$\mathrm{SDSC}\le 1$.

Combining both bounds, we conclude that $0 \le \mathrm{SDSC} \le 1$.
\end{proof}

\end{document}